# Design, Integration, and Evaluation of a Dual-Arm Robotic System for High Throughput Tissue Sampling from Potato Tubers

L.G. Divyanth[1,2], Syed Usama Bin Sabir[3], Divya Rathore[3], Lav R. Khot[1,2], Chakradhar Mattupalli[4], Manoj Karkee[1,2,3, *]

[1] Center for Precision & Automated Agricultural Systems, Washington State University, Prosser, WA 99350, USA

[2] Department of Biological Systems Engineering, Washington State University, Pullman, WA 99163, USA

[3] Department of Biological & Environmental Engineering, Cornell University, Ithaca, NY 14850, USA

[4] Department of Plant Pathology, Washington State University, Mount Vernon, WA 98273, USA

* Corresponding author: Manoj Karkee, PhD (manoj.karkee@wsu.edu)

**Abstract**

Manual tissue extraction from potato tubers for molecular detection of pathogens is highly laborious. Therefore, this study developed a machine-vision guided dual-arm coordinated inline robotic system that integrates tuber grasping and tissue sampling mechanisms. Functionally, tubers get transported on a conveyor system that halts when a YOLO11-based vision system identifies a tuber entering the workspace of a tuber gripping mechanism. A one-prismatic (P)-degree-of-freedom (DoF) arm equipped with a gripping end-effector securely grasps the tuber and positions it for sampling. The second robotic system, a 3-P-DoF Cartesian manipulator with a biopsy punch-based end-effector then performs tissue extraction for downstream molecular detection of pathogens. Using this system, tissue sampling was performed at specific location on a tuber (eyes or stolon scar), detected and localized by YOLOv10-enabled vision system. The tissue-sampling is a four-stage process: insertion of the end-effector into the tuber, rotation of the biopsy punch for tissue detachment from the tuber, biopsy punch extraction, and deposition of the tissue core onto designated collection location. This process was guided by the vision-based control scheme of the Cartesian manipulator. Results showed an average positional error of 1.84 mm at reaching the sampling location along the tuber surface, and an average depth deviation of 1.79 mm from the intended 7.00 mm punching depth. The overall success rate for tissue core extraction and deposition onto the designated collection location was 81.5%, with an average sampling cycle time of 10.4 s. The total cost of the system components was under $1,900, demonstrating the system's potential as a cost-effective alternative to labor-intensive manual tissue sampling. Future work will focus on optimizing the system for simultaneous sampling at multi-sites from a single tuber and broader testing at commercial operations.

## 1. Introduction

Potato is a valuable specialty crop grown in the United States (US), with an annual farm gate value of approximately $5 billion (USDA-NASS, 2024). Unlike other major crops such as corn, soybean, and wheat, which are propagated through true botanical seeds, potato is propagated vegetatively. This propagation method increases the risk of pathogen transmission across growing seasons. To address this challenge, there is a need to use high-throughput molecular-based approaches that are both sensitive and specific to target pathogens. Such a high-throughput workflow which involved sampling tissues from potato tubers and transferring them onto Whatman FTA Plantsaver® Cards (FTA) was developed by Ingram et al. (personal communication).

A crucial step in this workflow involves the manual extraction of tissue cores from specific locations on the tuber surface, including the eyes and stolon scar, where the probability of detecting pathogens such as potato virus Y (PVY) is greater (Schumpp et al., 2021). This process is performed using a biopsy punch, and the extracted samples are subsequently deposited onto FTA cards for downstream nucleic acid extractions and molecular detection of pathogens (Figure 1). However, manual tissue sampling process is laborious, requiring repetitive hand movements. For instance, about 8.5 person-hours are required for sampling 400 tubers. This poses scalability challenges if the potato industry in the U.S. were to adopt this approach. Therefore, automating the tuber sampling process is crucial and is the focus of this study.

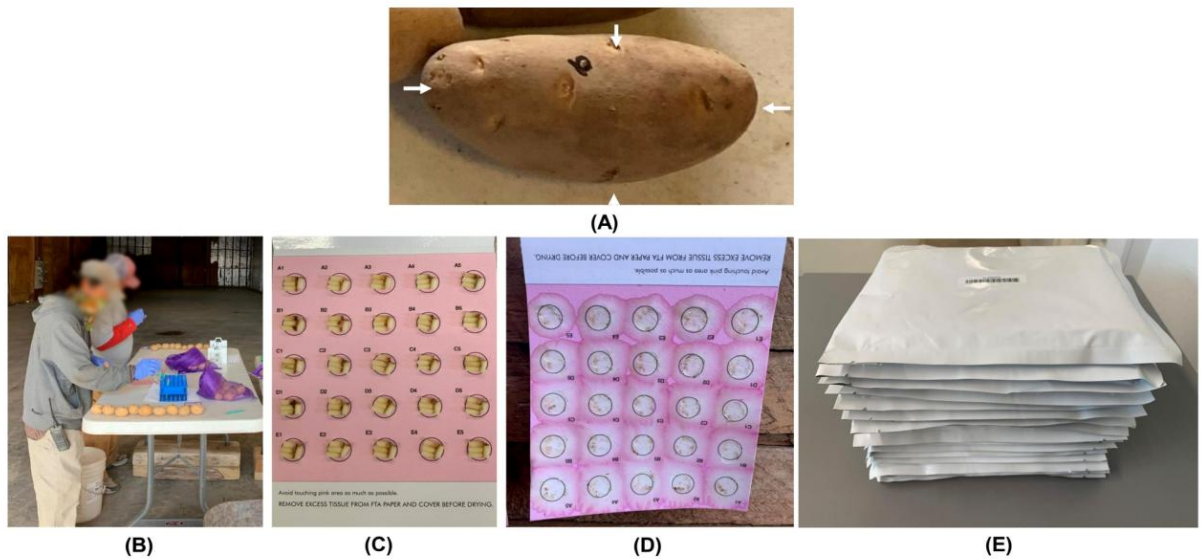

**Figure 1:** Potato tuber tissue sampling workflow (Divyanth et al., 2024). (A) White arrows indicate the targeted sampling locations on a tuber; (B) Manual extraction of tissue cores and placement onto FTA cards; (C) An FTA card carrying tissue cores from 25 tubers, with four samples per tuber; (D) Nucleic acids from the tissue cores are transferred onto the FTA card using a mechanical press; (E) FTA cards enclosed envelopes prepared for transport to a laboratory for PCR-based pathogen detection.

Advanced computer vision and robotic techniques provide can provide a promising framework for automated tissue sampling by integrating perception, manipulation, and specialized end-effectors/tools into a unified workflow (Oliveira et al., 2021; Thakur et al., 2023; Wakchaure et al., 2023; Zhang et al., 2020). Vision-guided robotic systems are particularly well-suited, as they have the capability to detect and localize tubers, securely grasp them, identify sampling sites, and precisely control the manipulator and end-effector for efficient tissue extraction. Furthermore, their compatibility with conveyor-based workflows (Mainali & Li, 2025; Yang et al., 2024) enables streamlined, repeatable operations, effectively mitigating the limitations of manual sampling. Despite the clear need, research efforts for robotic tissue sampling remain unexplored.

A typical robotic tissue sampling system must integrate four components: (1) an end-effector for extracting tissue samples; (2) a robotic manipulator for precise positioning and movement of the end-effector; (3) a tuber grasping mechanism to securely hold the tuber during sampling; and (4) a conveyor-based transport system for systematically delivering tubers to the sampling workspace. The design and integration of these components must be tailored to the operational environment to maximize efficiency, ensure precision, and minimize potential limitations.

End-effectors play a crucial role in robotic operations, enabling critical tasks such as object manipulation, grasping, and material extraction (Cepolina & Razzoli, 2024; Vrochidou et al., 2022). In agricultural applications, end-effectors have been extensively studied for tasks like tuber cutting (Huang et al., 2025; Singhpoo et al., 2024), transplanting (Almanzor et al., 2023; Liu et al., 2021), fruit harvesting (Li & Liu, 2023; Silwal et al., 2017; Zhang et al., 2023), tree pruning (You et al., 2023; Zahid et al., 2021), and plant phenotyping applications (Atefi et al., 2021), demonstrating innovative approaches to handling delicate biological materials. In the medical field, biopsy extraction tools have been widely explored for minimally invasive procedures (Biswas et al., 2023; Dagnino & Kundrat, 2024). Nonetheless, challenges remain in adapting existing robotic end-effectors for tissue sampling from potato tubers. Conventional grippers lack precision required for delicate or intricate procedures, making them unsuitable for tissue extraction.

A robotic manipulator is crucial for ensuring precision and repeatability in tissue sampling tasks. Various manipulator configurations, including Selective Compliance Assembly Robot Arms (SCARA), cylindrical, polar, articulated, Delta, and Cartesian systems, have been explored across different industries, each offering distinct operational advantages (Jin & Han, 2024; Zahedi et al., 2023). These systems differ in their Degrees of Freedom (DoF), workspace geometry, and kinematic structure, influencing key factors such as repeatability and scalability. For tissue sampling applications, high precision and repeatability are paramount, necessitating the selection and development of an effective manipulator with robust control and accuracy. Moreover, to address challenges posed by the irregular shapes and tendency of tubers to move or roll while performing tissue sampling, a dedicated grasping mechanism is essential alongside the tissue sampling component. Furthermore, the system must be incorporated with a conveyor-based transport system to enable a high-throughput and streamlined automated sampling operation.

Machine vision plays a pivotal role in robotic tissue sampling as it enables precise tuber detection, localization of sampling points, and the formulation of an effective control strategy. Previous studies have demonstrated the importance of perception-driven control in robotic applications such as tuber cutting (Huang et al., 2025; Singhpoo et al., 2024), fruit harvesting (Silwal et al., 2017; Zhang et al., 2021), and weed management (Evert et al., 2006; Zhao et al., 2025). In the context of this study, machine vision is essential for: (i) detecting incoming tubers to regulate the conveyor belt and ensure accurate positioning within the robot workspace, and (ii) identifying and localizing key sampling sites on a tuber, such as eyes and the stolon scar, to guide the manipulator in positioning the tissue sampling end-effector precisely.

Current research in machine-vision systems primarily focuses on detecting tubers and eyes in 2D RGB images, particularly for robotic tuber cutting applications (Gu et al., 2024; Huang et al., 2025; Liu et al., 2024). Traditional image-based methods, which rely on color, shape, and texture features (Mallahi et al., 2010; Liu et al., 2021; Tang et al., 2023), often struggle with variability in real-world conditions. In contrast, deep-learning techniques (Lecun et al., 2015), particularly convolutional neural networks (CNNs), automatically extract relevant features from imagery, resulting in more adaptable and less biased models. Recent studies have demonstrated the effectiveness of deep learning for the identification of eyes (Gu et al., 2024; Huang et al., 2025; Liu et al., 2024). Consequently, deep-learning-based perception modules will be explored in this study for tuber and sampling site detection, forming the basis of a control strategy for robotic tissue sampling.

In this study, we focus on developing an integrated robotic system for tissue sampling from specific regions of the tuber surface, such as the eyes and stolon scar, to enhance the efficiency of molecular pathogen detection workflow. The specific study objectives were to: i) design, develop, and evaluate a robotic high-throughput tissue sampling system integrating a specialized sampling end-effector, a manipulator, and a tuber grasping mechanism within a conveyor-based workflow; ii) investigate a machine-vision-based control scheme for coordinated operation between the tuber grasping mechanism, tissue sampling mechanism, and conveyor flow.

## 2. Materials and methods

### 2.1. Overview

The robotic tissue sampling system, illustrated in Fig. 2, consists of four key hardware modules: a biopsy-punch-based end-effector, a customized 3-DoF Cartesian manipulator, a tuber grasping mechanism, and a conveyor belt. An Intel® RealSense™ RGB-Depth camera (Intel Corporation, U.S.) mounted on the end-effector serves as the system's primary vision sensor. The system is fully integrated using the Robot Operating System (ROS 2 Humble Hawksbill, Open Robotics, U.S.) which facilitates communication and control among the modules. Machine vision tasks and overall system integration within the ROS framework are executed on an Alienware 17 R3 laptop (Dell Technologies, U.S.) equipped with an NVIDIA GeForce GTX 1070 GPU (NVIDIA Corporation,

U.S.), enabling high-performance processing. The camera and communication devices are interfaced with this laptop to support real-time data acquisition and processing.

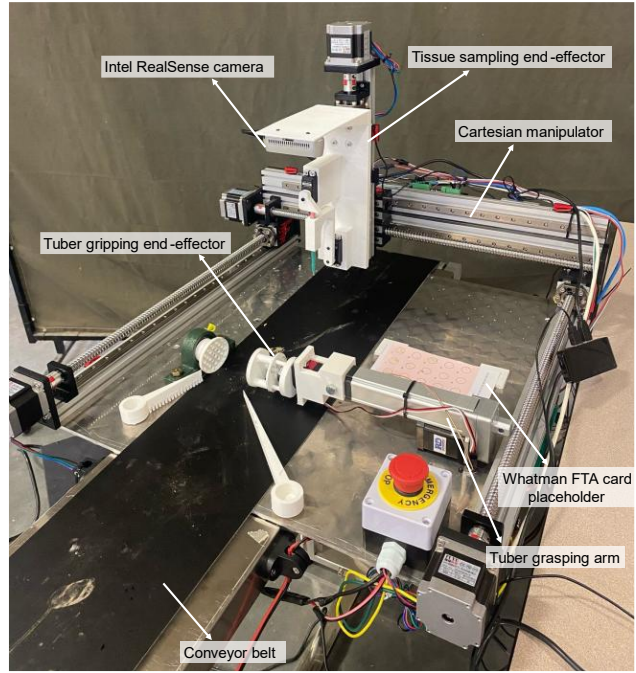

**Figure 2:** An overview of the robotic tissue sampling system components. The system includes a conveyor-based workflow for tuber transportation, a tuber grasping mechanism, and a camera-integrated tissue sampling end-effector mounted on a Cartesian manipulator.

The overall workflow of the robotic tissue sampling process is illustrated in Fig. 3. Initially, the camera is positioned at its home location, mounted on the tissue sampling end-effector, which is positioned at the highest point in the Cartesian manipulator workspace, directly above the midpoint of the conveyor belt. This placement maximizes the camera field of view and ensures optimal detection of incoming tubers. At this stage, a deep-learning-based tuber detection model detects the tubers and halts the conveyor when a tuber enters and aligns within the grasping arm's workspace. The grasping mechanism consists of a gripping end-effector mounted on a linear actuator oriented orthogonally to the conveyor flow. This configuration provides a prismatic DoF, enabling the actuator to clamp the tuber against a stationary gripper, effectively securing it between the two grippers. Once the tuber is secured, a second deep-learning-based detection model, the eyes and stolon scar detection model, identifies the sampling site on the tuber and guides the Cartesian manipulator to precisely position the end-effector for tissue extraction. After extracting the tissue core, the manipulator transports the end-effector to a designated deposition site, where the sample is placed on an FTA card. Simultaneously, the grasping mechanism releases the tuber. Finally, the manipulator returns the end-effector to its home position, and the conveyor resumes operation for sampling the next tuber.

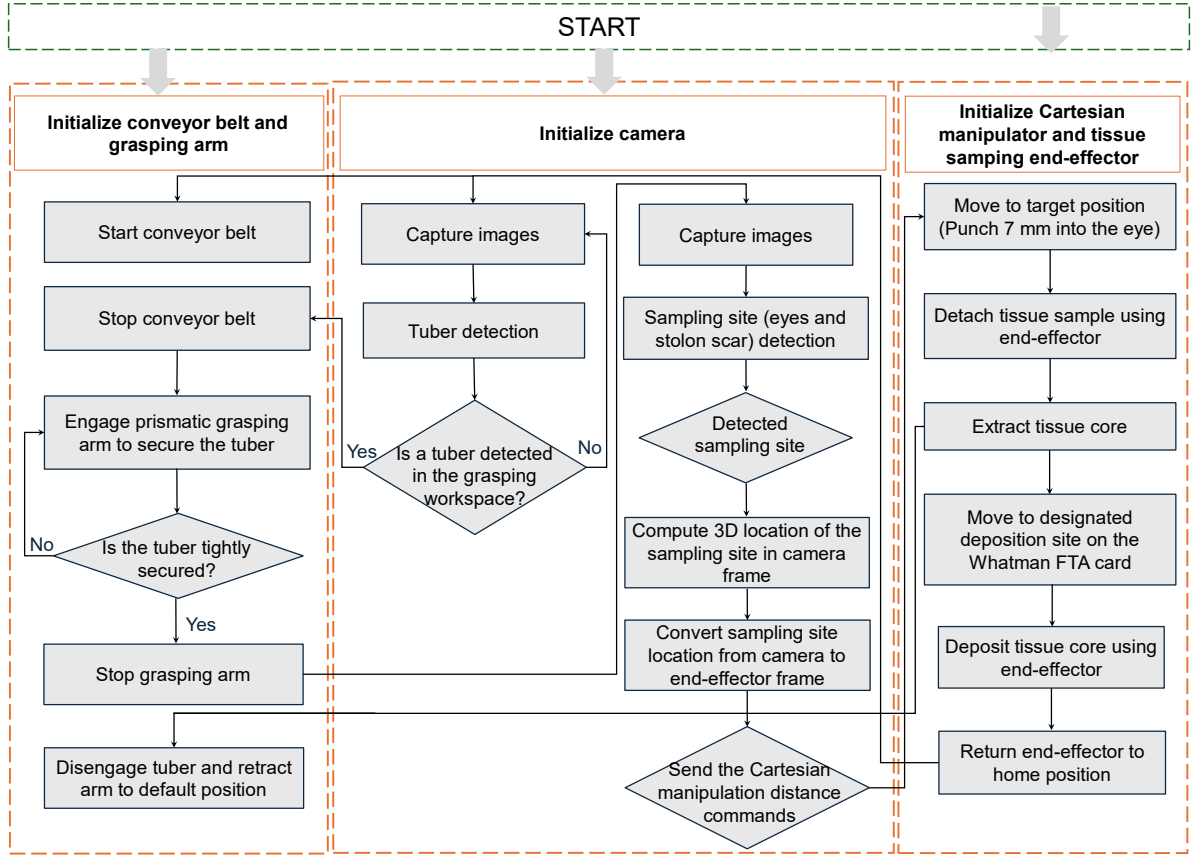

**Figure 3:** Flowchart of the robotic tuber tissue sampling process. The system operation begins with the initialization of the camera and conveyor belt to capture images and transport the tubers into the robot's workspace.

## 2.2. Tissue sampling end-effector

A biopsy punch will be used for manual tissue extraction from tubers, as illustrated in Fig. 4a. The procedure for tissue sampling involves inserting the biopsy punch into the tuber skin using its sharp-edged aluminum hollow cylindrical blade, followed by an angular extraction to detach the tissue core (Fig. 4b), ensuring that the tissue does not remain attached to the tuber. Upon withdrawal, the extracted tissue core is expelled from the punch through the piston mechanism (Fig. 4a (iii)), and a spring-loaded plunger inside the handle aids in resetting the piston to its original position following tissue deposition. Due to the proven efficiency and reliability in extracting tissue cores, it serves as the basis for developing a biopsy punch-based end-effector for the proposed robotic system for tissue sampling from tubers.

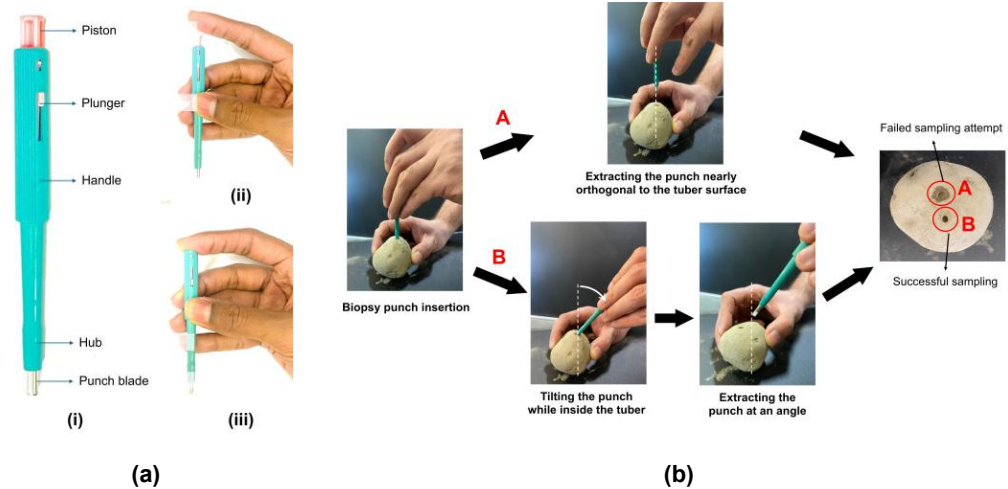

**Figure 4:** (a) (i) Illustration of the biopsy punch, a round-tipped cutting tool designed for extracting tissue samples from deeper layers of the tuber for diagnostic purposes, (ii) the punch is inserted into the tuber with its piston in non-actuated position, and (iii) piston in its actuated position for depositing the extracted tissue core; (b) a visual demonstration of the manual tissue extraction process, highlighting the method used to detach the tissue core efficiently.

The robotic end-effector for tissue sampling is designed to insert biopsy punch into the tuber skin, extract tissue cores, and release the extracted cores onto an FTA card. Additionally, the end-effector incorporates a camera to enable robot vision, essential for identifying the tuber and the sampling sites. Fig. 5 illustrates the end-effector consisting of a biopsy punch, actuating motors, and an RGB-D camera. The biopsy punch features a 3.25 mm diameter cylindrical blade with a length of 7.00 mm, allowing it to penetrate the tuber skin and extract tissue cores up to a depth of 7 mm. Notably, the blade is the only component of the end-effector that directly contacts the tuber.

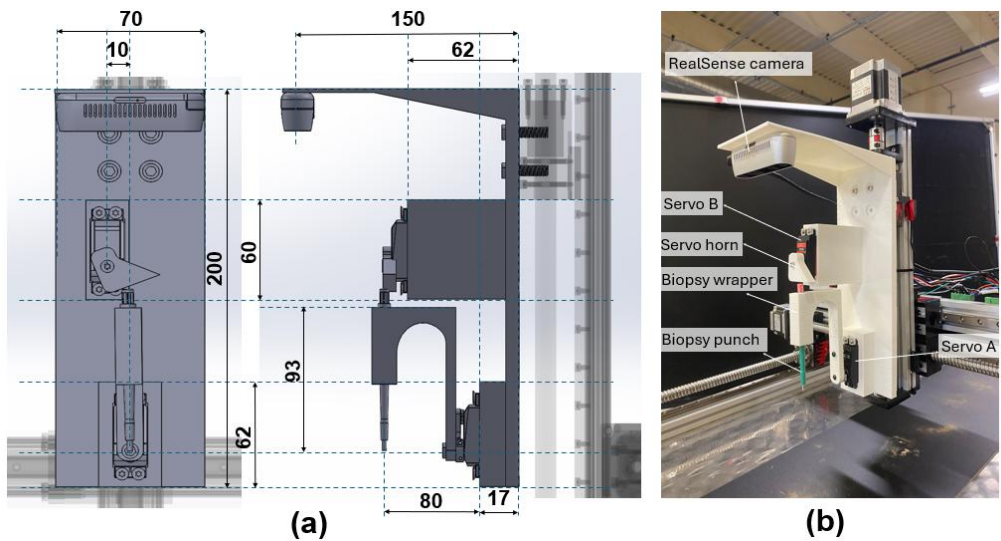

**Figure 5**: (a) A detailed 3D model of the tissue sampling end-effector with the dimensions (in mm); and (b) an assembled view of the end-effector, showcasing all key components.

A tilting motion of the biopsy punch was achieved by integrating the biopsy punch with a servo motor (called servo A), which is connected to the punch via a biopsy wrapper mounted to the motor's output shaft (Fig. 5b), providing a pitch (rotational) DoF for the biopsy punch. After the punch penetrates the tuber, the servo tilts it along the axis passing through its output shaft and the biopsy blade tip, facilitating effective tissue core detachment (Fig. 6a–c). Servo A is a 180° motor that delivers a maximum torque of 20 kg·cm, providing the necessary actuation force for tissue detachment.

A preliminary analysis was conducted to determine the optimal tilt angle, $\theta_p$, necessary for maximizing the tissue detachment success. The study examined $\theta_p$ at 20°, 30°, 40°, 50°, 60°, and 70°, revealing that highest accuracy was obtained with $\theta_p \geq 40°$. Based on these findings, $\theta_p = 40°$ was selected as the optimal angle for tissue extraction.

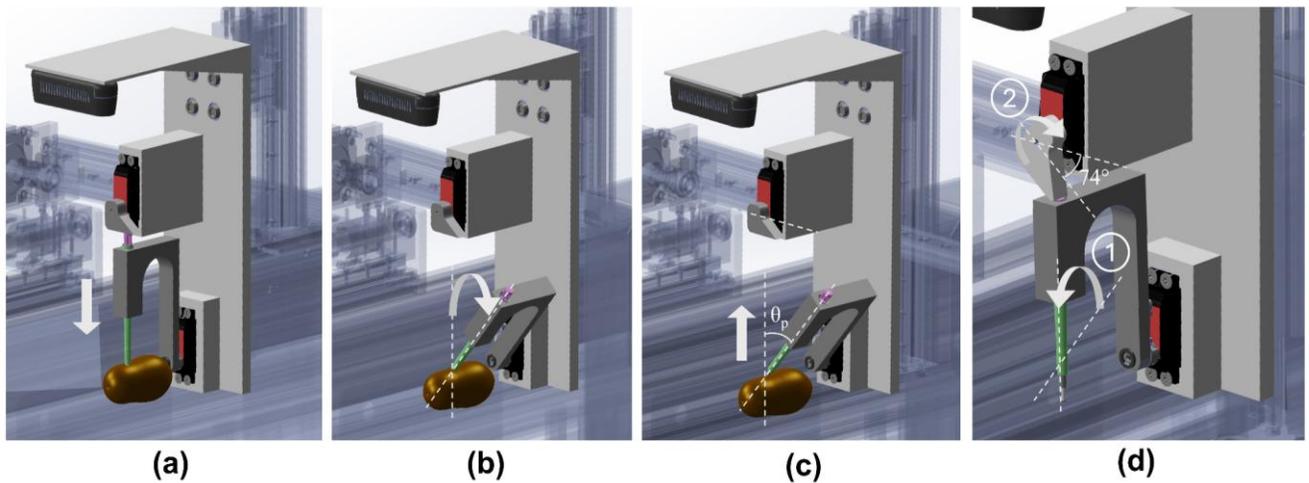

**Figure 6:** The sequence of key steps involved in the end-effector's operation during tissue sampling: (a) inserting the biopsy punch into the tuber; (b) pitch rotation of the punch to aid tissue detachment; (c) extracting the tissue core, and (d) depositing the tissue core.

A second servo motor (servo B) controls the piston within the biopsy punch to release the extracted tissue core. The design allows servo B to actuate the piston through a motor horn, enabling the controlled expulsion of the tissue core from the punch blade (Fig. 6d). To fully extend the piston, the motor horn rotates 74°.

The end-effector camera was positioned facing downward, observing the tubers on the conveyor belt as they moved into the manipulator workspace. It is mounted 200 mm above the end-effector base and 150 mm away from its mounting point on the manipulator, ensuring the tubers stay within the optimal depth-sensing range for imaging, while also preventing obstruction by the biopsy punch. Sufficient clearance was maintained between servo A and the biopsy punch (80 mm) (Fig. 5a) to ensure that the motor does not collide and disturb the tuber when the end-effector approaches the tuber and performs sampling. The dual-servo end-effector design enables efficient biopsy punch operation for reliable tissue core extraction and release, using open-loop feedforward control, without the need for additional sensors.

All structural components of the end-effector were fabricated using a Replicator 2X 3D printer (MakerBot Industries, U.S.), which reduced the overall mass of the device to ~0.51 kg and minimized the need for fasteners, simplifying the assembly.

## 2.3. Cartesian manipulator for tissue sampling

For tissue sampling tasks that demand high precision and repeatability, Cartesian manipulators are highly effective. Unlike more flexible systems, such as articulated or SCARA arms, which offer a broader range of motion, Cartesian systems operate along three linear axes (X, Y, and Z), providing straightforward control and superior accuracy (Arad et al., 2020; Barnett et al., 2020; Chen et al., 2021; Zahid et al., 2020). Therefore, a three degree-of-freedom (DoF) Cartesian manipulator was designed and assembled to guide the tissue sampling end-effector to the desired locations (Fig. 7). This configuration maximizes both the precision required for tissue extraction and the simplicity of the design.

The tissue sampling end-effector was mounted on a linear actuator that controls the z-axis, enabling vertical movements of the end-effector for punching into the tuber and pulling it back for extracting the sampled tissue core. The z-axis actuator was positioned orthogonally to a second linear actuator forming the y-axis. Additionally, two parallel linear actuators at the base define the x-axis, oriented orthogonally to both the y- and z-axes, allowing for coordinated movement in all three dimensions (Fig. 7). The manipulator was constructed using stepper motor-driven ball screw actuators with linear guide rails (FSL40 linear guide stage actuator with a NEMA 23 stepper motor, Chengdu Fuyu Technology Ltd., China), capable of supporting a maximum vertical load of 25 kg and providing a maximum torque of approximately 0.95 N·m. The selection of these actuators was informed by an empirical study, in which a human operator manually inserted the biopsy punch into the tuber using a single-finger vertical force. A force sensor placed on the operator's finger measured the peak distal normal force exerted at the point of punch insertion. These measurements were used to determine the required actuation force for the robotic system, ensuring that the selected actuators could deliver the necessary insertion force.

The stepper motor-driven actuators provide precise rotational control, enabling highly accurate and repeatable motion along predefined axes. This accuracy ensures consistent and precise manipulation throughout the sampling process. The z-axis actuator has a full stroke of 30 cm, but its effective travel range is limited to 10 cm by IR-based transistor limit switches at both ends of the actuator rails, preventing the end-effector from hitting the base and ensuring a safe movement range. Similarly, the x- and y-axis actuators each have full stroke ranges of 60 cm × 60 cm, but their effective workspace is constrained to 38 cm × 46 cm. This limitation ensures a compact operating area and safe operation, while accommodating both the grasping mechanism, which holds the tuber, and the FTA card for tissue deposition within the workspace (Fig. 7).

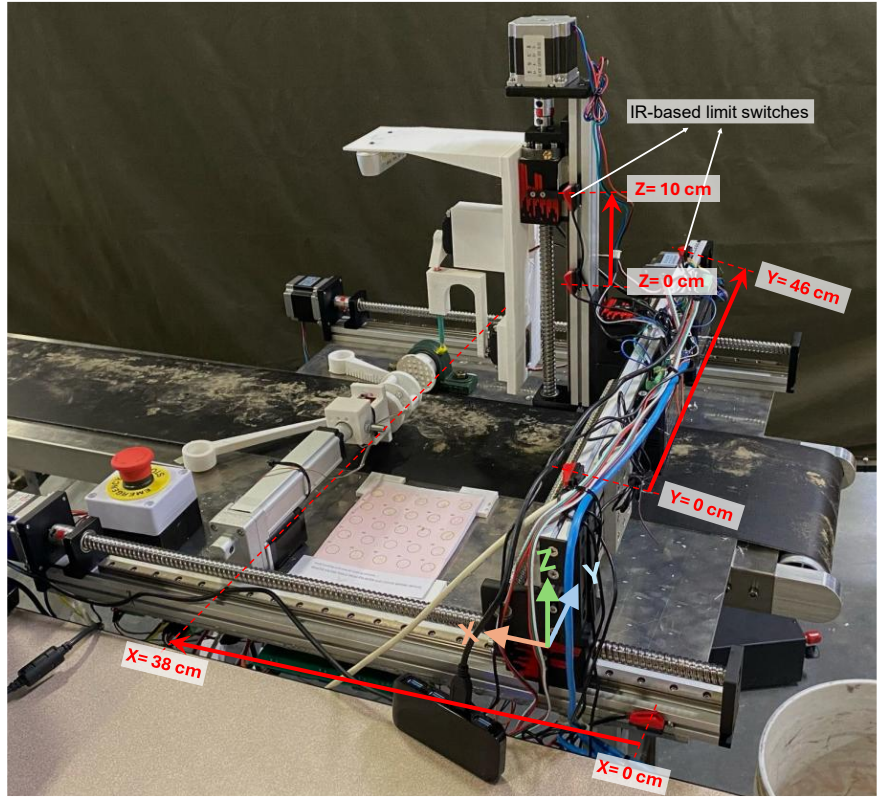

**Figure 7:** Illustration of the Cartesian manipulator's axes system and their travel range. The travel limits were set using IR-based transistors acting as limit switches.

### 2.4. Tuber grasping mechanism

To address the challenge posed by the irregular shapes and tendency of potato tubers to move or roll while performing tissue sampling, a dedicated grasping mechanism was designed to hold the tubers in place before tissue sampling process was executed.

The grasping mechanism consisted of a linear actuator-based arm equipped with a gripping end-effector (Fig. 8). The end-effector comprised two main components: a sliding gripper and a fixed base. The sliding gripper used a spiked contact surface to enhance friction with the tuber, while its opposite side incorporated three cylindrical sleeves acting as sliding bushings. The circular contact surface of the gripper had a diameter of 60 mm, designed to securely grasp tubers typically ranging from 35 to 60 mm in length. The fixed base contained three cylindrical guides functioning as sliders that fit into the sleeves of the sliding gripper. These components were mechanically coupled via compression springs (spring constant, $K = 100$ N/m), which contracted when a perpendicular force was exerted on the gripping surface.

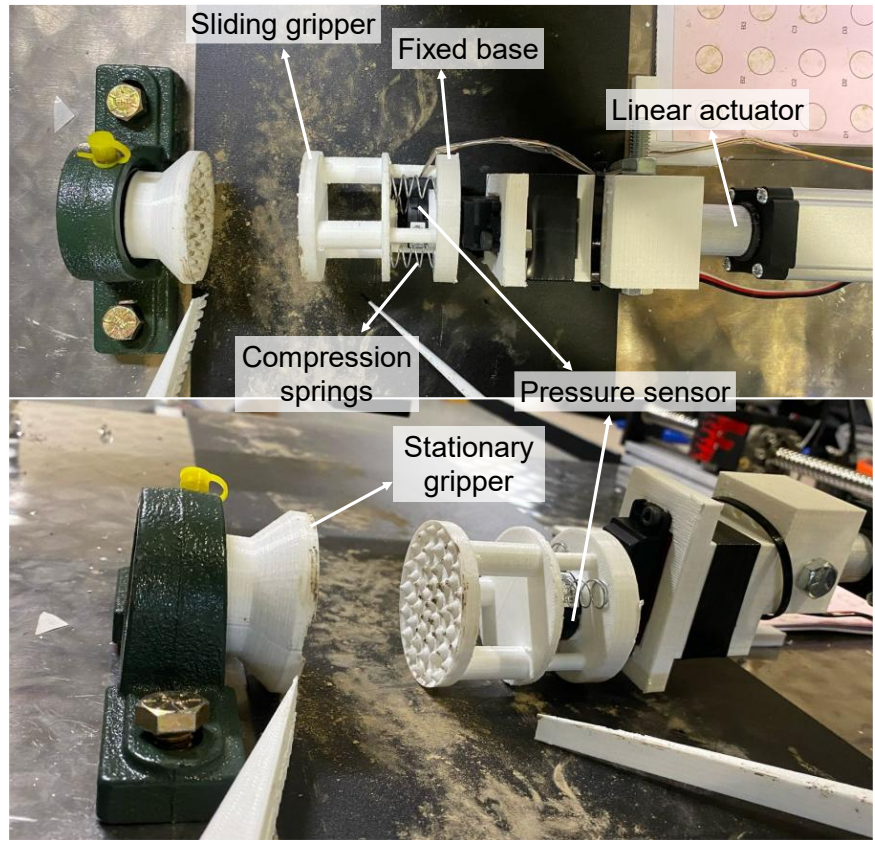

**Figure 8:** Assembled view of the tuber grasping mechanism.

The linear actuator was powered by a stepper motor (RobotDig Shanghai, China), providing a prismatic DoF with a maximum stroke length of 14 cm and a linear velocity of 4 mm/s. The arm was mounted perpendicular to the tuber's transportation direction on a conveyor system. On the opposite side, a rigidly mounted stationary gripper was positioned concentrically with the grasping arm's gripper. When a tuber aligned between the two grippers, the linear actuator engaged, pushing the tuber against the stationary gripper to secure it in place. The compression spring introduced a preload force, ensuring firm and adaptive clamping of the tuber.

A calibrated force/pressure sensor (TE Connectivity Measurement Specialties, Inc., U.S.; model number: FX292X-040B-0100-L) was incorporated on the fixed base of the gripping end-effector, positioned to detect contact with the sliding gripper. As the tuber was clamped, the sliding gripper compressed the springs and exerted force on the sensor. The linear actuation was disengaged when the force sensor registered 0.8 N, a threshold optimized through a preliminary experiment to securely hold the tuber during tissue sampling without allowing the spikes to pierce its surface.

### 2.5. Mechanical system integration for inline tissue sampling

To streamline the workflow and enable high-throughput sampling, the system was integrated into a conveyor-based transport mechanism. The conveyor system facilitated the continuous movement of tubers, allowing the grasping and tissue sampling mechanisms to operate efficiently in an automated, inline process. The conveyor belt was driven by a DC motor and had a width of 253 mm, a thickness of 2 mm, and a length of 2100 mm. The conveyor speed was set to 20 mm/s, a value optimized for synchronization with the system's communication speed. Guides were installed along the conveyor to align the tubers into a single-file arrangement for sequential transportation to the workspace. The Cartesian manipulator was mounted with its x-axis aligned parallel to the conveyor's motion, while the grasping arm's prismatic movement was perpendicular to the conveyor flow. At the start of each tissue sampling cycle, the camera was positioned at a home location such that the grasped tuber remained within its field of view.

## 2.6. Robot perception system

The robotic tissue sampling system relies on RGB-D imagery for tuber and sampling site detection and localization. To achieve this, an Intel® RealSense™ D435i RGB-D camera was employed as the primary vision sensor, providing both color and depth information. Two deep-learning-based object detection models were developed for the system: one for detecting tubers being transported on the conveyor to the grasping workspace and another for identifying sampling locations (eyes and stolon scar) on the tuber surface.

An image dataset (called tuber image dataset) was collected using the camera mounted on the end-effector, while positioned at its home location in the Cartesian manipulator. This setup, integrated with the conveyor belt, represented the expected operational environment and camera's field of view during the robotic tissue sampling operation (Fig. 9a). A russet potato cultivar (Clearwater russet) was selected for the experiments in this study. While the tubers were transported on the conveyor belt, a total of 260 RGB images from about 126 tubers were captured using the RealSense camera.

### 2.6.1. Tuber detection model

For developing the tuber detection model, the tuber image dataset was annotated for tubers using Roboflow software (Roboflow Inc., U.S.) (Fig. 9b). Of the total 260 images, 220 images were used for training the model, while the remaining 40 images were used for validation. To enhance model robustness, various data augmentation techniques were applied to the training set. These techniques included geometric augmentations, such as flipping along the x- and y-axes, translation (±25%), scaling (±25%), mosaic composition, and rotations (±30°), as well as image intensity-based augmentations, such as brightness adjustments (coefficients: 0.7, 0.8, 0.9, 1.1, 1.2, 1.3), HSV adjustments (H = 0.015; S = 0.7; V = 0.4), and Gaussian blurring (standard deviation = 1.5). Ground-truth bounding box information was preserved and replicated for all augmented images.

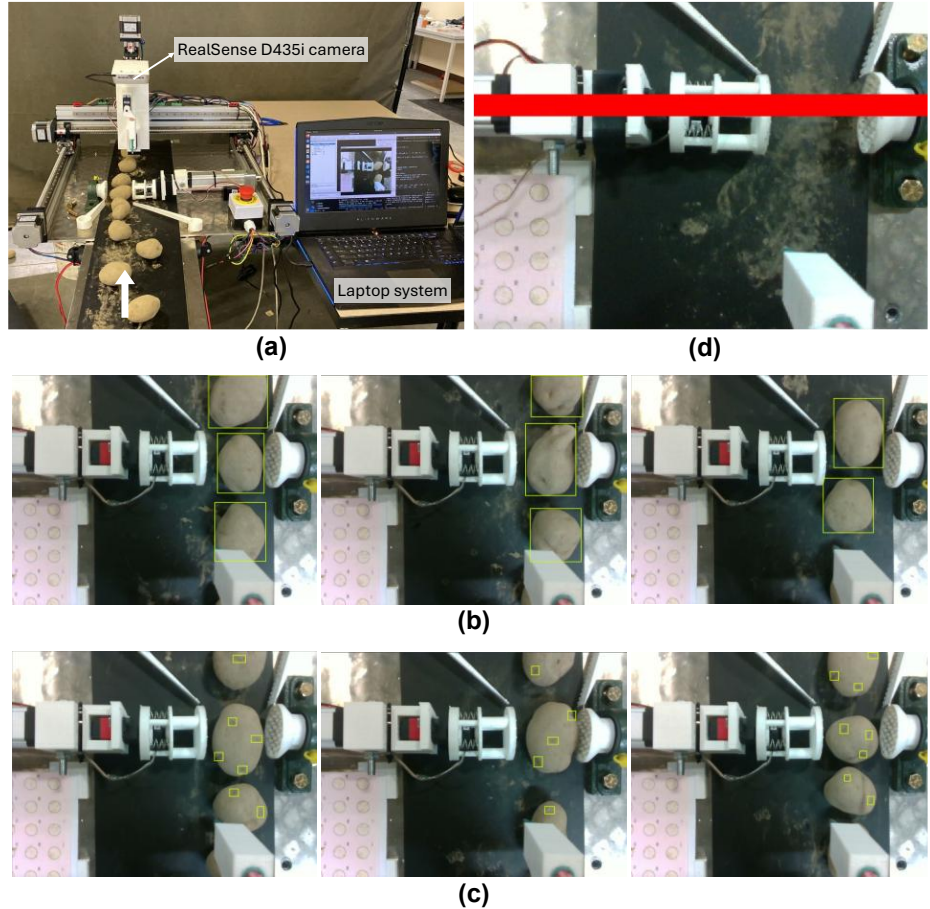

**Figure 9:** (a) The image acquisition setup used for collecting the tuber image dataset; (b) Example images of tubers with bounding box annotations, utilized for developing the tuber detection model; (c) Example images used for developing the eyes and stolon scar detection model; (d) The reference region (highlighted in red) defined in the RealSense-captured images, indicating the tuber grasping workspace.

The YOLO family of deep neural networks represent state-of-the-art object detection algorithms with demonstrated success in diverse applications (Badgujar et al., 2024; Dang et al., 2023; Wei et al., 2024). Hence, the YOLO algorithm, specifically the latest YOLO11n, was selected for developing the tuber detection model. Image annotations in Roboflow software were exported in a YOLO-compatible label format. The model was trained using transfer learning by fine-tuning weights pre-trained on the COCO dataset. Input image dimensions were standardized to $640 \times 640$ pixels to align with the YOLO11 architecture. Training was conducted for 100 epochs with a mini-batch size of 16 images using the PyTorch framework. The training process was performed on an Alienware 15 R3 system equipped with an NVIDIA GeForce GTX 1060 GPU running on Ubuntu operating system.

During the tissue sampling operation, while the camera-mounted end-effector remained in the home position, the YOLO11n-based tuber detection model processed the RGB image stream to detect the tubers on the conveyor belt. In the captured images, a rectangular region corresponding to the workspace of the grasping arm was defined to monitor tuber positioning (Fig. 9d). When the centroid of a detected tuber's bounding box aligned with this reference region, the conveyor belt was halted, ensuring appropriate alignment for grasping. The detailed control scheme is discussed in section 2.6.

### *2.6.2. Eyes and stolon scar detection and sampling point localization*

After the tuber is grasped, a separate deep-learning-based object detection model was used to identify specific sampling locations, such as the eyes and stolon scars on potato tubers. Divyanth et al. (2024) evaluated the performance of YOLO-based deep-learning models for efficient eye and stolon scar detection, addressing challenges such as variability in eye characteristics across different tuber cultivars. This study assessed nineteen architectures of the YOLO versions (YOLOv5, YOLOv6, YOLOv7, YOLOv8, YOLOv9, YOLOv10, YOLO11) for combined eye and stolon scar detection, providing valuable insights for selecting the optimal model for the robotic tuber sampling system. The YOLOv10m-based eye detection model demonstrated superior performance, achieving an mAP@0.5 of 0.914, with a precision of 0.903 and a recall rate of 0.815, while maintaining an average inference time of 92ms. Details of the dataset and training methodology used to develop this model can be found in (Divyanth et al., 2024). Based on these findings, the YOLOv10m model was selected as the baseline for identifying sampling locations. The tuber image dataset was annotated for eyes and stolon scars (Fig. 9c), and the YOLOv10m-based model was further fine-tuned on this dataset using transfer learning to adapt to the specific environmental and operational conditions of the tissue sampling system. The training process followed the same approach as the tuber detection model, utilizing the same training-validation data split, data augmentation, and training parameters.

During tissue sampling, once a tuber was grasped, an algorithm based on the eyes and stolon scar detection model was used to identify the most reliable sampling location. Before executing the eyes and stolon scar detection model, the bounding box coordinates of the grasped tuber (from the tuber detection model) were used to isolate its region in the RGB image, effectively filtering out other tubers. This approach also improved detection accuracy by reducing false positives caused by mud and dust in the workspace, which sometimes exhibited features resembling tuber eyes. After detecting the potential sampling locations on the tuber using the detection model, selecting a site closer to the centroid $(x_{tC}, y_{tC})$ of the tuber's bounding box resulted in higher-quality tissue extraction. Sampling near the tuber's boundaries often led to reduced tissue retention in the punch and increased localization errors due to surface curvature. Therefore, the model selected the most reliable sampling location by choosing the detection closest to the centroid of the tuber, identified by comparing the Euclidean distances between the centroid pixel coordinates of the detected sampling sites' bounding boxes and the centroid of the tuber.

To determine the 3D coordinates of the sampling site, the centroid pixel coordinates of the selected bounding box $(x_{sC}, y_{sC})$ was used to retrieve the corresponding depth value $Z_{sC}$ from the depth channel provided by the RealSense camera:

$$Z_{sC} = Depth(x_{sC}, y_{sC}) \tag{1}$$

Using this depth value and the centroid's pixel coordinates, the Cartesian coordinates $(X_{sC}, Y_{sC}, Z_{sC})$ of the location in camera's frame were computed via back-projection:

$$X_{sC} = \frac{(x_{sC} - c_x) \cdot Z_{sC}}{f_x} \tag{2}$$

$$Y_{sC} = \frac{(y_{sC} - c_y) \cdot Z_{sC}}{f_y} \tag{3}$$

where $f_x$ and $f_y$ are the focal lengths in the x- and y-directions, respectively, defining how the camera projects real-world points onto the image plane, and $c_x$ and $c_y$ are the optical center coordinates in pixels. Notably, the pixel coordinates $(x_{sC}, y_{sC})$ were corrected for lens distortion inherent in the RealSense camera using the radial and tangential distortion coefficients provided as the camera's intrinsic parameters (Herrera et al., 2012). The computed 3D coordinates were used to guide the end-effector to the sampling site (motion control described in Section 2.6).

### 2.7. Control system design

This section outlines a control strategy that integrates machine vision algorithms with the mechanical components of the robotic system, ensuring the coordinated operation of the tuber grasping and tissue sampling mechanisms within a conveyor-based workflow.

#### 2.7.1. Control system architecture

A control interface was developed using Python and C++ within a ROS framework. The complete hardware architecture and the communication links between the various peripherals used in this study are illustrated in Figure 10. The system architecture consists of a laptop (Alienware 17 R3, Dell Technologies, U.S.) acting as the central communication hub, interfacing with the RGB-D camera and the microcontroller (Arduino Mega 2560, Arduino Inc., Italy). A custom-designed printed circuit board (PCB), developed using EasyPCB software (EasyPCBUSA, U.S.), was integrated with the microcontroller to facilitate signal and power distribution. The system incorporates four stepper motors: three dedicated to independent prismatic motion along the x-, y-, and z-axes of the Cartesian manipulator and one for the prismatic movement of the tuber grasping arm, as discussed before. Each stepper motor is driven by a dedicated stepper motor driver, controlled using pulse-width modulation (PWM) signals from the microcontroller. For safety and operational constraints, six limit switches are positioned at the ends of the linear actuators of the Cartesian manipulator.

The tissue sampling end-effector is actuated by two servos, where servo A maintains the biopsy punch's orientation, keeping it vertical during punching and tissue release while tilting it to a specific angle ($\theta_p$) for tissue detachment, and servo B actuates the piston mechanism of the biopsy punch, toggling between an open state (natural) and a closed state for tissue release. The force sensor integrated in the tuber gripping end-effector monitors the clamping force, transmitting its readings through an HX711 amplifier module (SparkFun Electronics, U.S.) to the microcontroller.

The conveyor system, powered by a DC motor, is controlled via a solid-state relay (SSR) through the microcontroller, enabling a pause-resume operation in response to machine vision feedback. All the electrical components were powered by a high-current switching DC power supply (BK Precision 1694, BK Precision, U.S.), with required step-down (12/24 VDC) converters used to drive the stepper motor linear actuators, servo motors, and the limit switches.

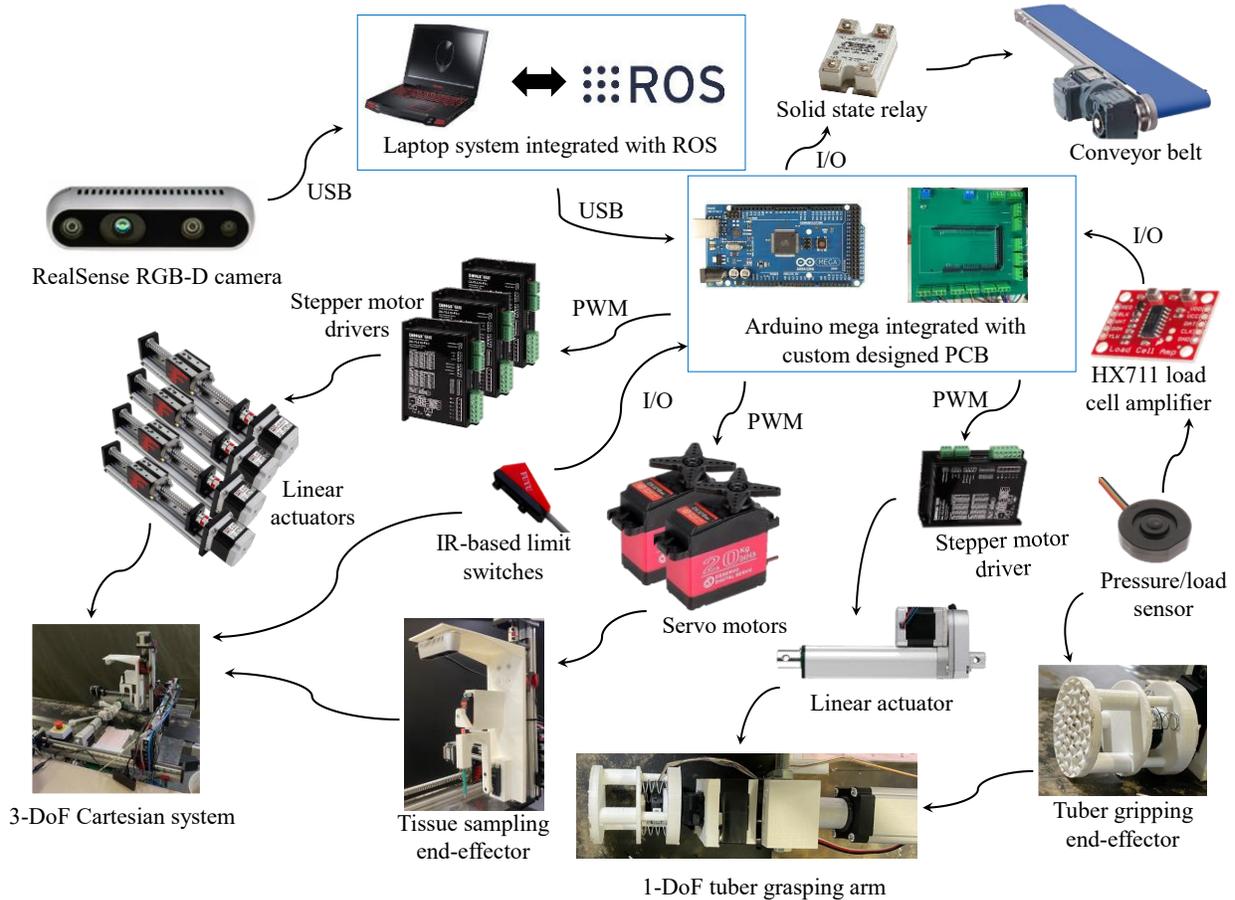

**Figure 10:** The developed control system of the tissue sampling robot showing its key components.

### 2.7.2. Machine-vision-based motion control

The coordinate systems defined for the tissue sampling robot are illustrated in Figure 11. At the start of each sampling cycle, the origin of the end-effector frame is located at the home position—located at the highest possible point in the Cartesian workspace, directly above the midpoint of the conveyor belt ($x = 120\ mm, y = 330\ mm, z = 100\ mm$). The tuber detection algorithm computed the centroids ($x_{t_k}, y_{t_k}$) for each of the $n$ detected tuber bounding boxes (where, $k$ ranges from 1 to $n$). In the captured RGB images of size ($x_i = 640, y_i = 480$) pixels, the grasping workspace is defined by pixels ($x_g \in [0, x_i]$, $y_g \in [y_{g_l} = 165, y_{g_h} = 195]$) (Fig. 9d). When a detected tuber's centroid aligns within this region (i.e., $y_{t_k} \in [165, 195]$), the controller halts the conveyor and actuates the grasping arm's prismatic movement until the force sensor registers the threshold force.

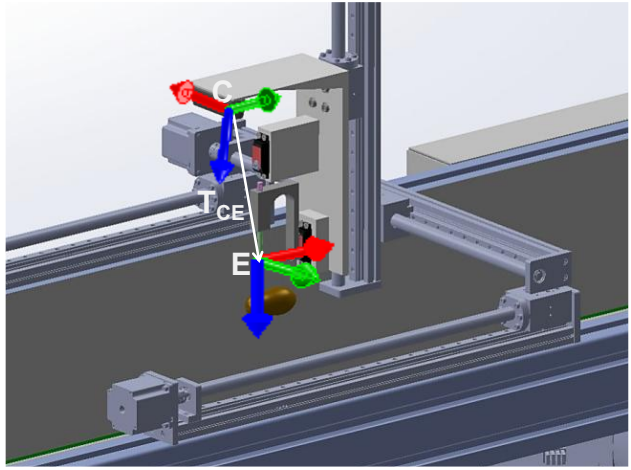

**Figure 11:** The camera frame ($C$) and end-effector frame ($E$) coordinate systems (all are right-hand coordinate frames) in the tissue sampling robot, and transformation from camera frame to end-effector frame ($T_{CE}$).

Once the tuber was grasped, the eyes and stolon scar detection algorithm was executed. The algorithm computed the 3D centroid coordinates ($X_{sC}, Y_{sC}, Z_{sC}$) of the bounding box corresponding to the selected sampling site, based on equations (1–3), in the camera frame. However, to guide the end-effector using the Cartesian manipulator, these coordinates needed to be transformed into the end-effector's frame. The transformation matrix between the end-effector and camera frames, $T_{CE}$, was generated using hand-eye calibration, following the procedure outlined by Hu et al. (2022). These centroid coordinates were then transformed into the end-effector's coordinate frame ($X_{sE}, Y_{sE}, Z_{sE}$) using the transformation matrix $T_{CE}$.

$$\begin{bmatrix} X_{sE} \\ Y_{sE} \\ Z_{sE} \\ 1 \end{bmatrix} = T_{CE} \cdot \begin{bmatrix} X_{sC} \\ Y_{sC} \\ Z_{sC} \\ 1 \end{bmatrix} \qquad (4)$$

where,

$$T_{CE} = \begin{bmatrix} 0.0003 & 0.9998 & 0 & -51.447 \\ -0.9998 & 0.0003 & 0 & 18.095 \\ 0 & 0 & 1 & 179.142 \\ 0 & 0 & 0 & 1 \end{bmatrix} \qquad (5)$$

The manipulator then moves directly to the target ($X_{sE}, Y_{sE}$) coordinates without any movement along the z-axis, positioning itself above the sampling site. Once at the target, it moves vertically downward to $Z_{sE} + 7\ mm$, ensuring optimal biopsy punch insertion into the tuber. After the punch is inserted into the tuber surface, servo A tilts the biopsy punch to the desired angle, $\theta_p$, to facilitate tissue detachment.

A holder is mounted in the X-Y plane at the Z-axis origin of the Cartesian workspace to hold the FTA card, which contains 25 designated tissue collection circles arranged in a 5 × 5 grid (Fig. 12). Tissue cores from each tuber are iteratively deposited into these circles. Once all circles are filled with tissue cores, the card can be taken for diagnostic purposes and replaced with a new card in the placeholder. For the $(n, m)$-th circle, where $n, m \in [1, 5]$, the center coordinates relative to the initial home position of the end-effector are defined as $(X_{c_{n,m}}, Y_{c_{n,m}}, Z_{c_{n,m}})_H$. To ensure precise deposition, the end-effector releases the tissue core 10 mm above the center of the designated circle, $(X_{c_{n,m}}, Y_{c_{n,m}}, Z_{c_{n,m}} - 10\ mm)_H$. This 10 mm vertical offset guarantees accurate tissue core placement within the designated circle, without the risk of displacing the circles.

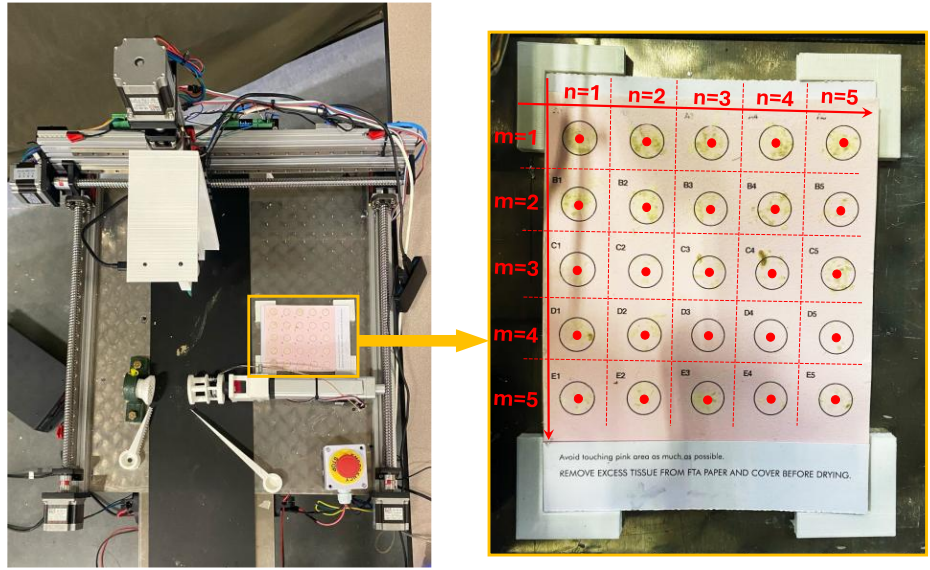

**Figure 12:** Illustration of the Cartesian workspace setup, depicting the Whatman FTA card with the sample collection circles.

Therefore, following tissue detachment, the manipulator uses a step counter variable defined for the stepper motors of the Cartesian axes, and the predefined sample deposition location $(X_{c_{n,m}}, Y_{c_{n,m}}, Z_{c_{n,m}} - 10\ mm)_H$, to guide the end-effector to the deposition site. First, the end-effector retracts 40 mm vertically to safely disengage from the tuber. It then moves to the deposition location, where the tissue core is released. The steps are expressed as:

$$(X_{sE}, Y_{sE}, Z_{sE} + 7\ mm)_H \rightarrow (X_{sE}, Y_{sE}, Z_{sE} - 33\ \text{mm})_H \rightarrow (X_{c_{n,m}}, Y_{c_{n,m}}, Z_{c_{n,m}} - 10\ mm)_H$$

where, $(X, Y, Z)_H$ refers to the coordinate location from the home position.

Upon arrival, servo A repositions the biopsy punch to vertical orientation, while servo B activates the piston mechanism to release the tissue core. After release, servo B resets to its open state, while the Cartesian manipulator returns the end-effector to the home position. At the same time, the grasping arm retracts to release the tuber, thus completing the tissue sampling cycle. Once the end-effector reaches the home position, the conveyor belt resumes movement. The tuber detection model is activated after t = 1s, ensuring the sampled tuber has cleared the grasping workspace (since the belt moves at 20 mm/s). The vision-controlled conveyor then positions the next tuber for grasping, initiating the next sampling cycle.

The actuators of the Cartesian manipulator follow a constant velocity profile, maintaining a speed of 50 mm/s throughout the entire movement. The manipulator operates under open-loop control, ensuring precise motion execution without step loss. To mitigate any risk of step loss, the step counter is recalibrated at the commencement of each sampling cycle, when the manipulator returns to its home position.

## 2.8. Experiment design for system evaluation

A comprehensive evaluation of the proposed robotic tissue-sampling system was conducted at the Center for Precision & Automated Agricultural Systems (CPAAS) at Washington State University. The experiment utilized 81 russet potato tubers, distinct from those used for developing the tuber image dataset. Due to the limited length of the experimental conveyor belt, the tubers were fed in nine separate batches. Each batch was loaded by a human operator only after the system had completed tissue sampling from the previous batch. The evaluation aimed to assess the performance of the critical components of the system, including the tuber and sampling location detection models, the reliability of the tuber grasping mechanism, the accuracy of the machine-vision-based Cartesian motion control system, the effectiveness of the tissue-sampling end-effector, and the overall efficiency of the system in terms of task execution time.

To evaluate the deep learning models, two datasets were recorded. The first dataset, which was used to assess the performance of the tuber detection model, consisted of 81 images captured by the camera as the tubers were transported on the conveyor belt. These images were annotated for ground-truth by marking the locations of the tubers, and the model was evaluated by comparing its predictions to these ground-truth annotations. Performance metrics, including mean Average Precision at an Intersection over Union (IoU) threshold of 0.5 (mAP@0.5), precision, recall, and inference speed, were used to quantify the model's effectiveness (Rainio et al., 2024). Similarly, for the eyes and stolon scar detection model, images were captured of the 81 test tubers as they were grasped by the robotic system, and ground-truth annotations were made to identify the locations of the eyes and stolon scars. The model was compared with this ground-truth and evaluated using the same metrics as for the tuber detection model.

The performance of the machine-vision-based open-loop motion control system for Cartesian manipulators was assessed by analyzing the end-effector's positioning during the manipulation process, particularly focusing on lateral deviation on the tuber surface and penetration depth accuracy. Further, success rates for the tissue extraction and deposition tasks were recorded to quantify the performance of the end-effector. Finally, a quantitative analysis of the extracted tissue cores was performed by measuring their length using a vernier caliper.

A critical aspect of the evaluation was the measurement of task execution time. The time taken for each major task in the workflow was recorded, with the tasks categorized into vision-system processing, grasping mechanism operation, Cartesian manipulation tasks, and tissue sampling end-effector operation. The RealSense camera operated at 6 frames per second to provide the necessary time for processing each image.

## 3. Results and discussion

### 3.1. Machine-vision system

As discussed in the methods section, a total of 81 tissue sampling cycles were completed. The vision system was utilized twice per cycle: once at the beginning for tuber detection on the conveyor belt, and again for sampling location detection and localization. To minimize sensor deviation and errors, the median value of the 3D location of the sampling point was computed using 10 consecutive RGB and depth frames. Image acquisition was performed while the end-effector remained stationary during both instances, preventing potential errors caused by motion blur. Example image frames illustrating the detections at both stages of the vision system are shown in Fig. 13.

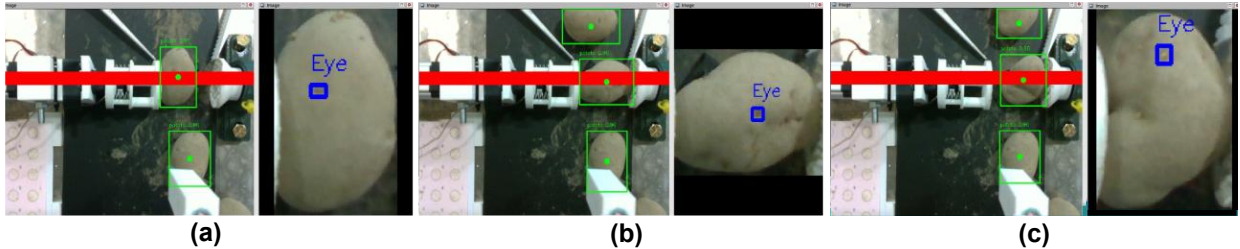

(a)          (b)          (c)

**Figure 13:** Sample images illustrating results of the tuber, as well as eyes and stolon scar detection models. Green bounding boxes indicate tuber detections, while the green dot represents the centroid of the detected tuber. Once a tuber's centroid aligns with the grasping workspace (red line), the tuber is grasped, and the eyes and stolon scar detection model is activated to identify potential sampling locations. Among the identified locations, the sampling site closest to the centroid of the tuber is selected for tissue sampling, as indicated by the blue bounding box.

The potato tuber detection model, based on the YOLO11n configuration, achieved an mAP@0.5 value of 0.985, while achieving perfect detection accuracy (precision and recall values of 1.000; or successful identification of all tubers) during the evaluation. The model demonstrated an average inference time of just 37 ms per image, demonstrating its capability to achieve both accurate detection and high computational speed. Given the relatively simple nature of the tuber detection task with minimal complexity, the smaller YOLO11n model was sufficient to meet the accuracy requirements.

On the other hand, the YOLOv10-based eyes and stolon scar detection model achieved an mAP@0.5 of 0.924, with a precision of 0.911, indicating that 91.1% of the detected sampling locations were correct, and a recall of 0.889, meaning that 88.9% of all actual sampling locations in the dataset were successfully detected. The model maintained an average inference time of 92 ms.

### 3.2. Performance of the tuber grasping mechanism

The conveyor system effectively transported and aligned the tubers within the grasping workspace, utilizing real-time machine vision feedback from the tuber detection model. As each tuber reached the grasping zone, the robotic arm executed a prismatic motion with an average displacement of 10.4 mm to securely grasp the tuber. The gripping mechanism, equipped with spiked contact surfaces, ensured sufficient friction for a stable grip, while force sensor feedback prevented the gripper from penetrating the tuber.

Tubers with irregular or asymmetrical shapes, including those with uneven surfaces or tapered ends, tended to roll unpredictably and move away from the grasping area during the gripper's attempt to position them. Tubers with high aspect ratios or uneven weight distributions, particularly those with smooth and asymmetrical surfaces, were especially prone to shifting direction and escaping the grasping workspace. The gripper's design featuring protruding tabs along the rims, helped minimize this issue to some extent. Out of the 81 tubers transported to the grasping workspace, 73 (90.1%) were successfully secured by the gripper. The remaining 8 tubers required human intervention to adjust their orientation for a successful grasp.

### 3.3. Performance of vision-guided Cartesian manipulation

The performance of the vision-guided motion control for the Cartesian manipulator was evaluated based on the accuracy of the end-effector's positioning as it navigated to the identified sampling locations on the tuber surface. Results (Fig. 14A) showed an average positional error of 1.84 mm along the tuber surface. Specifically, 25% of the sampling attempts achieved a positional error of less than 1 mm, while approximately 20% exhibited errors greater than 3 mm. It was observed that higher accuracy was achieved when the selected sampling location was near the centroid of the tuber, where the back-projection method used for localization benefited from more reliable depth information from the RealSense camera. In contrast, errors were more significant when the sampling locations were near the edges, as the depth distortion caused by the curvature negatively affected precision.

The vision-system occasionally computed a target penetration depth exceeding 7.00 mm, causing the biopsy punch to attempt deeper insertions than intended (17 out of 81 attempts). However, the larger cross-sectional diameter of the biopsy tool's hub (Fig. 4a (i)) prevented over-penetration by limiting the torque from the Cartesian manipulator's actuators, ensuring a consistent and effective insertion depth of 7.00 mm. When these cases were excluded (because the punching depth in those cases were always as expected), the average depth deviation was 1.79 mm less than the intended 7.00 mm punching depth (Fig. 14B). Conversely, sampling attempts near the edges of the tuber frequently resulted in insufficient penetration depth (deviation > 3 mm). This was primarily attributed to inaccurate depth estimation caused by curvature-induced distortions, significantly affecting tissue core integrity and overall sampling quality.

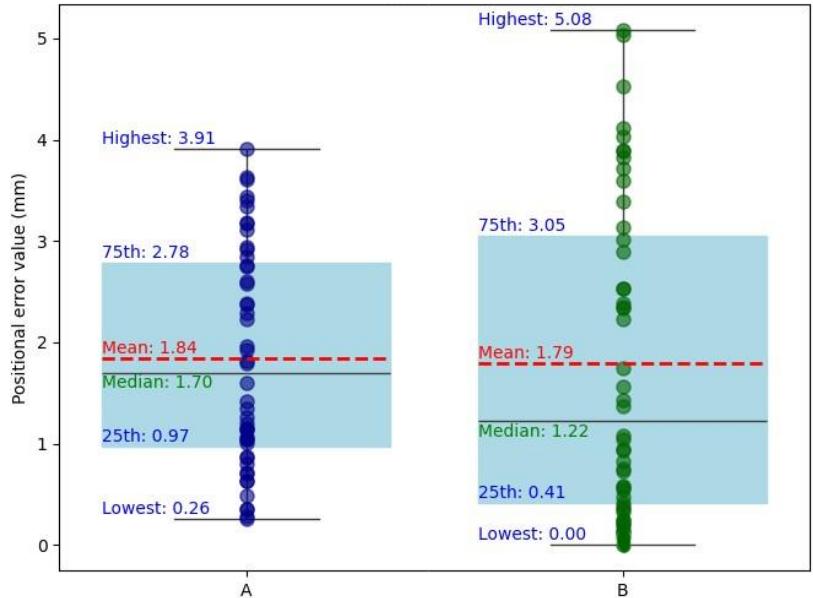

**Figure 14:** A box plot illustrating the distribution of the Cartesian manipulation deviation in reaching the desired (A) sampling point location on the tuber surface and (B) sampling depth.

### 3.3. Performance of the tissue-sampling end-effector

The motion control scheme effectively guided the biopsy punch to the identified sampling location, enabling tissue extraction and subsequent deposition at the designated site. The performance of the tissue sampling end-effector was evaluated through experiments assessing its efficiency at each stage of the sampling process. The overall success rate of the end-effector in extracting a tissue core and accurately depositing it within the designated tissue collection circle on the FTA card was 81.48% (66 out of 81 attempts). Figure 15 illustrates an example when the end-effector successfully performed the tissue sampling operation. The 15 unsuccessful sampling cycles were analyzed and categorized into three major failure modes:

1. Tissue detachment failure (5/15): This failure mode (33.33% of the failure) was primarily attributed to positional errors in the motion control system. If the biopsy punch failed to penetrate the tuber skin sufficiently (typically < 3 mm), the tissue mass inside the blade was inadequate for detachment. As a result, the tissue core remained loosely attached to the tuber and was not properly extracted.

2. Unsuccessful tissue deposition (8/15): This issue was caused by two primary factors:

   *Tissue core adhesion to the ejecting piston (4/15):* In some cases, the extracted tissue core adhered to the ejecting end of the piston, preventing it from falling out during deposition. When the piston retracted, the tissue core was pulled back into the biopsy blade, disrupting subsequent sampling cycles due to residual tissue inside the punch.

   *Hub interference with the tuber surface (4/15):* If the vision system inaccurately computed the depth of the sampling site (typically when error > 4 mm), it led to an improper positioning of the biopsy punch. When the computed sampling depth was deeper than the actual depth, the hub inadvertently contacted the tuber surface. Since the hub has a larger cross-sectional diameter than the blade, it could not penetrate the tuber. This miscalculation resulted in errors in step execution by the Cartesian manipulator. Over successive cycles, these accumulated errors led to the biopsy tool being positioned too close to or directly in contact with the FTA card during deposition, making it difficult for the tissue core to be properly released.

3. Tissue core carrying failure (2/15): In this failure mode, the vibrations in the overall system during the manipulator movement caused the extracted tissue core to become dislodged from the biopsy punch before reaching the deposition site.

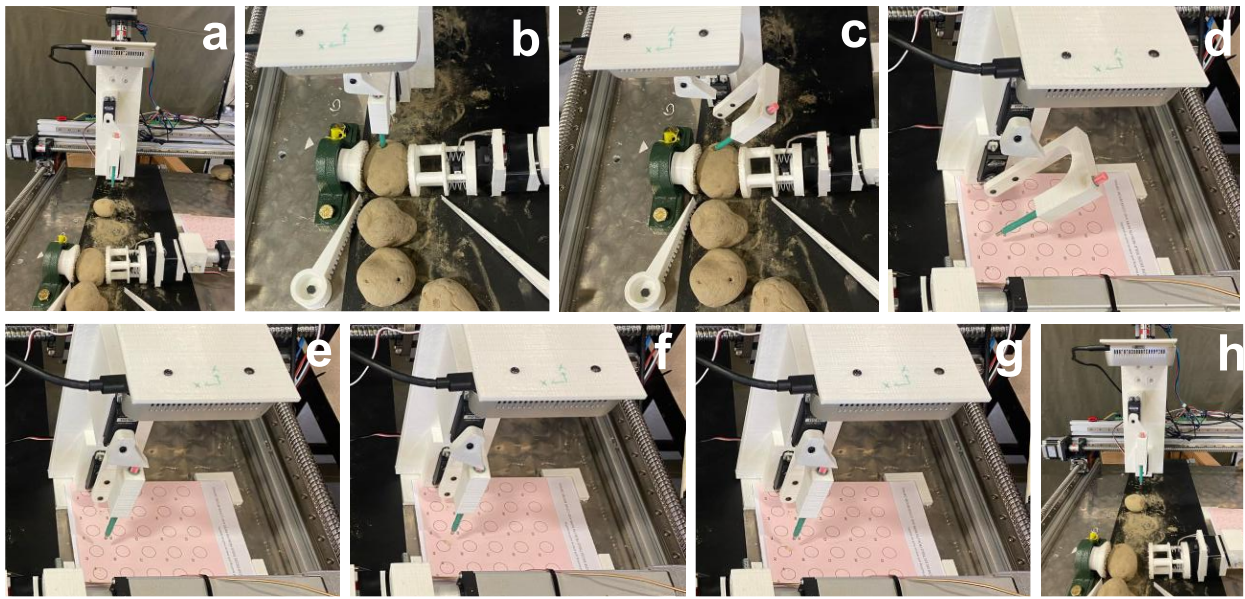

**Figure 15:** Snapshots of a tissue sampling cycle: (a) The tuber is grasped by the gripping mechanism, with the end-effector in the home position, (b) the manipulator guides the end-effector to the sampling location, enabling the insertion of the biopsy blade, (c) tissue is detached by rotating the biopsy punch, (d) the tissue is extracted by the end-effector through the vertical movement of the manipulator, (e) the manipulator guides the end-effector to the tissue deposition location, (f) the end-effector brings the biopsy punch to a vertical position, (g) tissue is expelled through the biopsy piston, actuated by the servo horn, (h) the end-effector is returned to the home position, preparing for the next tissue sampling cycle.

A post-inspection of the 76 successfully sampled tissue cores (except the 5 attempts that resulted in tissue detachment failure) revealed significant variations in length, a crucial factor influencing sample quality for serological analysis. Notably, the length of the extracted tissue cores did not always equal the punch insertion depth, as detachment did not consistently occur precisely at the end of the biopsy blade during rotation inside the tuber. The analysis showed that the tissue cores ranged in length from 2.81 mm to 6.92 mm, with a mean length of 5.84 mm (Fig. 16). Among the sampled cores, 61.84% (47/76) were ≥ 6 mm, 23.68% (18/76) measured between 4 mm and 6 mm, and 14.47% (11/76) measured < 4 mm. These findings suggest that while many samples met the required length, inconsistence persisted, highlighting opportunities for optimizing the biopsy punch design for higher volume of tissue core retention.

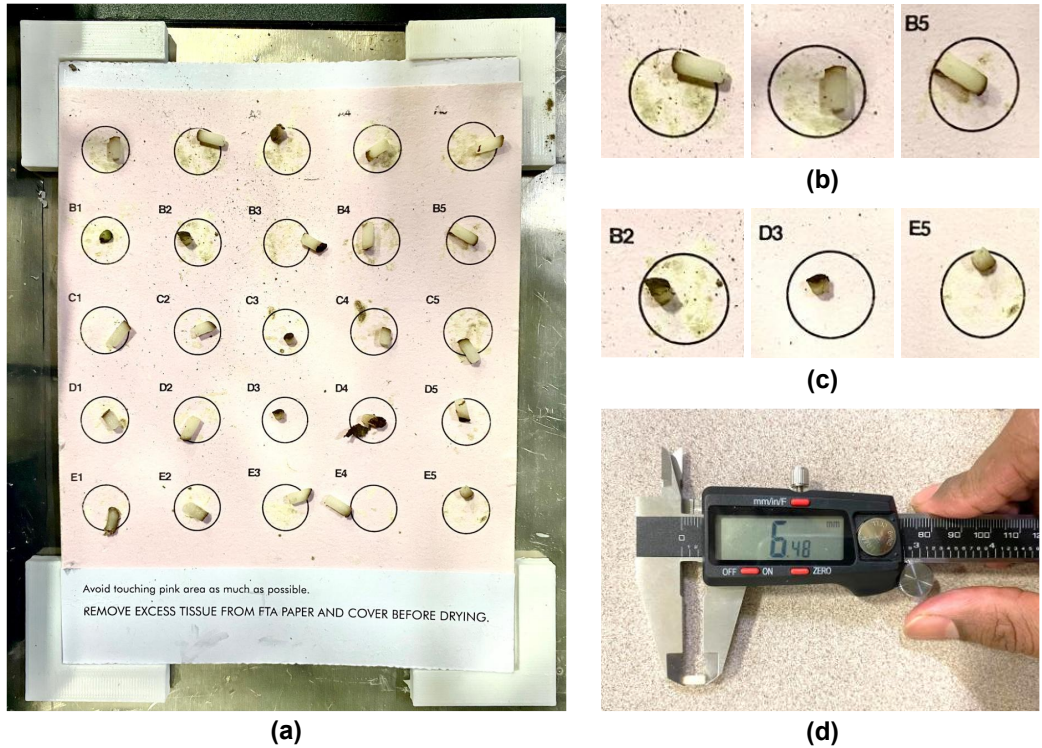

**Figure 16:** (a) Tissue cores systematically deposited onto the circles of a Whatman FTA card; (b) examples of tissue cores with sufficient length (> 6 mm); (c) examples of tissue cores with an unsatisfactory length (< 4 mm); and (d) tissue core lengths measured using a vernier caliper.

One of the primary factors influencing tissue core length was the penetration depth of the biopsy punch. Since the biopsy punch rotates about the end of its blade during tissue extraction, the tissue detaches around this point. If the punch failed to penetrate deep enough due to motion control inaccuracies, the extracted cores were shorter than intended. This issue was particularly pronounced when the punch encountered areas with high surface curvature, leading to inconsistent tissue retention across the biopsy blade's cross-section. Additionally, samples taken from the peripheral regions of the tuber, rather than centrally, were more likely to be shorter due to surface curvature effects, as mentioned before.

### 3.4. Task execution time analysis

As described earlier, the camera was operated at 6 frames per second, thus allowing about 167 ms per frame for image processing and computations. Since tuber detection was performed in real-time concurrently while the tubers were transported on the conveyor belt, it did not contribute to the overall cycle time. The effective vision-related processing time accounted for RGB and depth image acquisition, detection and localization of the sampling points (processed over 10 frames per cycle), and computation of Cartesian distances for robotic manipulation. On average (based on n = 81 sampling cycles), vision-related tasks require 2.1 s per cycle. The grasping mechanism required an average of 2.6 s to securely hold the tuber.

The duration of each Cartesian manipulation subtask was recorded in every sampling cycle. These tasks included approaching and inserting the end-effector into the sampling site (1.9 s), transporting the extracted tissue core to the designated collection area (2.2 s), and moving the end-effector back to home position (1.6 s). On average, the total time required for manipulation tasks was 5.7 s per cycle. End-effector operations that included biopsy punch rotation for tissue detachment and biopsy piston actuation for tissue release were open-loop commands set at 0.4 s and 0.6 s, respectively ($\times$ 2 for bringing the punch and piston to natural position). Therefore, including the average vision time of 2.1 s, the total cycle time required for sampling a single tissue core was 12.4 s. It should be noted that the reported cycle time does not include the time required for transporting a new tuber into the workspace via the conveyor belt, as the inter-tuber distances on the belt vary, introducing variability in overall cycle timing.

## 4. Implications for future work

The goal of this research was to develop and evaluate a robotic system for in-line tissue sampling of potato tubers. The system consisted of a dedicated tuber grasping mechanism, a tissue sampling end-effector, and a Cartesian manipulator, integrated with a machine-vision-based control scheme. The total cost for all mechanical hardware and vision system components was under $1,900, demonstrating potential as an alternative to labor-intensive manual tissue sampling. The key lessons learned from this research include the following themes:

- The proposed vision system and algorithms show the potential to robustly detect and localize the tubers and sampling sites, particularly given the challenges associated with detecting eyes on potato tubers (Divyanth et al., 2024). The positional error achieved during the robotic manipulation to the identified sampling site could be attributed to the RGB-D camera sensor error, minor inaccuracies in back-projecting the object locations from image-frame to the camera-frame, and frame transformation from the camera to end-effector. Calibration changes due to the heating of the camera could be another reason. Further improvement in camera-end-effector calibration methods needs to be considered (Pezzuolo et al., 2018; Wang et al., 2022; Yu et al., 2024).

- The biopsy punch-based end-effector demonstrated high effectiveness in tissue detachment and extraction, with no recorded failures attributed to these processes. The primary challenge observed during the tissue sampling cycle was in the deposition phase, where occasional inconsistencies in tissue release led to improper placement onto the designated collection area. This issue arose due to adhesion between the tissue core and the piston of the biopsy punch, preventing ejection. A potential improvement to mitigate this problem would be the integration of a vacuum-based suction and release mechanism (Gharakhani et al., 2022; Lu et al., 2022; Hua et al., 2025) within the biopsy punch, which could facilitate the reliable expulsion of tissue cores.

- The robotic system demonstrated promising potential for automated tissue sampling when tested on russet potato tubers. Hence, further research is necessary to validate its performance across various potato cultivars. An important area for improvement is refining the biopsy punch design to achieve precise penetration depths while minimizing step errors caused by the hub during insertion. Additionally, integrating an encoder-based feedback mechanism in the Cartesian system could enable real-time depth monitoring, ensuring consistent and accurate biopsy blade penetration.

- Most of the sampling attempts with higher errors and inconsistencies were observed in regions away from the tuber's center, as viewed by the machine vision system. This was primarily due to the curvature of the tuber and reduced depth estimation accuracy at the periphery (Schreiberhuber et al., 2019; Harkel et al., 2017). To address this issue, incorporating an additional rotational DoF in the grasping mechanism could enable reorientation of the tuber, ensuring that sampling points are positioned to the central region for improved accuracy. Additionally, future work should focus on extracting multiple tissue cores from a single tuber, including representative samples from the stolon scar and multiple eye locations. In fact, the same tuber rotation mechanism would enable sampling from multiple locations in each tuber, which would enhance the diagnostic process by providing a more comprehensive and representative tissue collection.

## 4. Conclusion

This study developed an inline, fully automated robotic tissue sampling system that will help with high-throughput molecular detection of pathogens from potato tubers. The system consists of a tuber grasping mechanism and a tissue sampling end-effector mounted on a Cartesian manipulator. A machine-vision-based control strategy was integrated with the overall system. Specifically, a YOLO11-based tuber detection model controls the conveyor belt to transport and align the tubers within the grasping workspace. Once positioned, the tuber is grasped using a 1-DoF prismatic arm and a gripping end-effector. A YOLOv10m model identifies the sampling locations, including the eyes and stolon scar, on the grasped tuber and guides the integrated tissue sampling end-effector and 3-DoF Cartesian manipulator to perform tissue sampling with open-loop control. The system successfully identified 88.9% of the sampling locations on the tubers, with an average localization time of 2.1 s per tuber. Of the 81 tissue sampling cycles attempted, 66 were successfully completed, with an average cycle time of 10.4 s. The study also highlighted potential areas for improvement, particularly in the design of the tissue-sampling end-effector, such as integrating a vacuum-based suction and releasing mechanism within the biopsy punch. Additionally, enhancing the grasping mechanism by incorporating an extra rotational DoF for reorienting the tuber could improve sampling accuracy. Furthermore, refining the control methods to enable multi-site tissue sampling from a single tuber would further enhance the system's efficiency and throughput.

**Acknowledgments**

This work was supported by the Washington State University – Emerging Research Initiative. We sincerely thank Dr. Mark Pavek and the potato grower cooperators for providing the tubers used for image acquisition and experimentation.